\relax
\pdfoutput=1
\documentclass{article}
\usepackage{arxiv}  
\usepackage{times}  
\usepackage{helvet} 
\usepackage{courier}  
\usepackage[hyphens]{url}  
\usepackage{graphicx} 
\usepackage{amsfonts}
\usepackage{amsmath}
\usepackage{cite}

\usepackage{caption}
\usepackage{subcaption}
\usepackage[switch]{lineno}  
 
\usepackage{graphicx}  
\title{Providing Actionable Feedback in Hiring Marketplaces using Generative Adversarial Networks}
\author{Daniel Nemirovsky, Nicolas Thiebaut, Ye Xu, Abhishek Gupta\\
Hired, Inc.\\ 
San Francisco, CA \\
\{daniel.nemirovsky, nicolas.thiebaut, ye.xu, abhishek.gupta\} @hired.com
}
\date{}

\begin{document}

\maketitle


\begin{abstract}
  Machine learning predictors have been increasingly applied in production settings, including in one of the world's largest hiring platforms, Hired, to provide a better candidate and recruiter experience. The ability to provide actionable feedback is desirable for candidates to improve their chances of achieving success in the marketplace. Until recently, however, methods aimed at providing actionable feedback have been limited in terms of \textit{realism} and \textit{latency}. In this work, we demonstrate how, by applying a newly introduced method based on Generative Adversarial Networks (GANs), we are able to overcome these limitations and provide actionable feedback in real-time to candidates in production settings. Our experimental results highlight the significant benefits of utilizing a GAN-based approach on our dataset relative to two other state-of-the-art approaches (including over 1000x latency gains). We also illustrate the potential impact of this approach in detail on two real candidate profile examples.
\end{abstract}

\section{Introduction \label{sec:intro}}
\noindent 
The prevalence of machine learning (ML) based predictive models has increased throughout various industries including recruitment platforms and marketplaces. Providing model interpretability and actionable feedback is increasingly viewed as an integral part of any production ML system.

\begin{figure}[!ht]
\centering
\includegraphics[width=0.5\columnwidth]{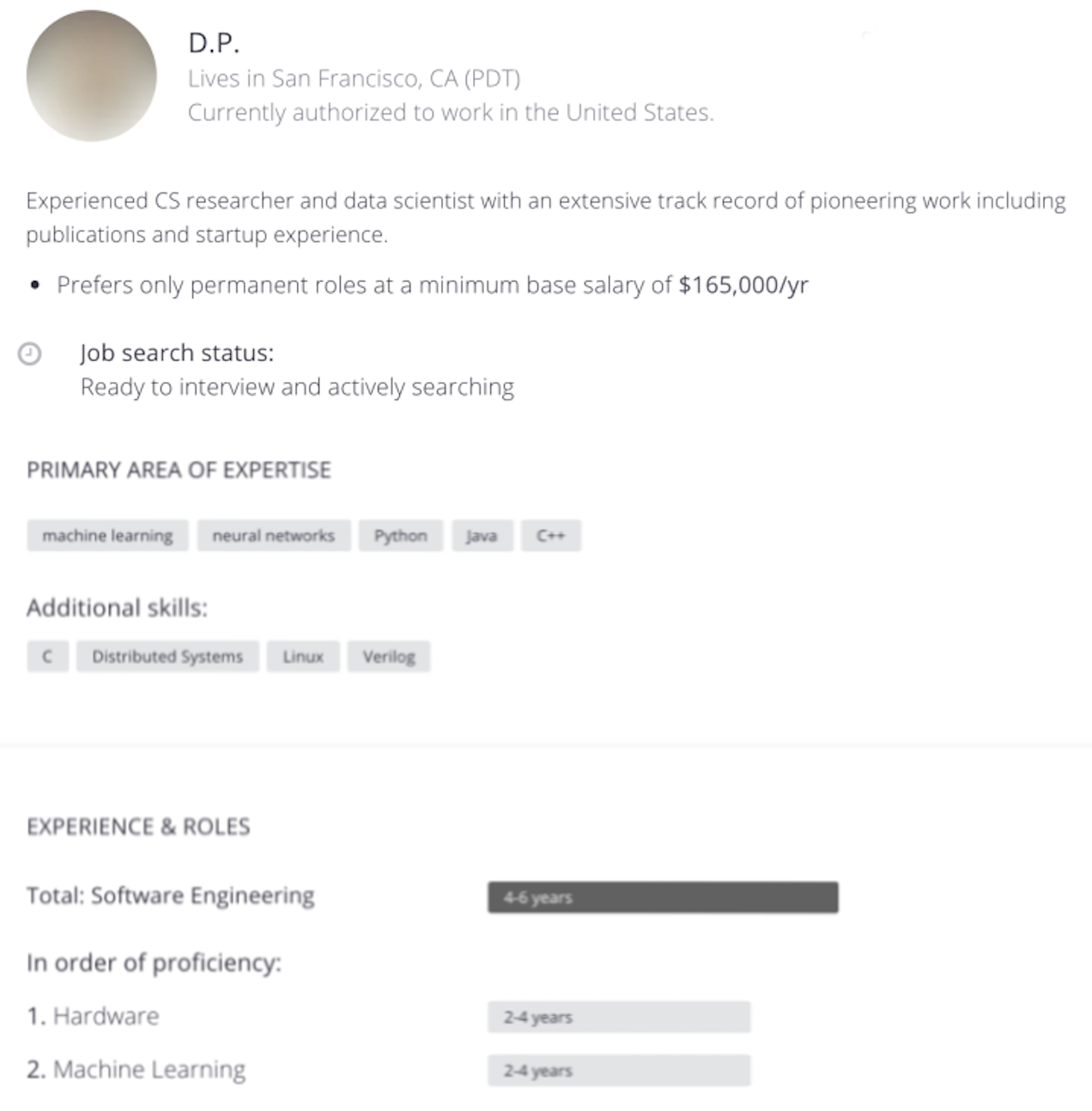}
\caption{An example of a candidate's profile (redacted) on the Hired marketplace (https://hired.com/).}
\label{fig:hired_profile_ex}
\end{figure}

At one of the world's largest hiring marketplaces, Hired, candidates fill out a profile, similar to a resume with structured data such as skills, work experience, preferred location, and salary. They then apply to be approved onto the marketplace for recruiters to discover and interview. An ML classifier is used to approve candidates on the marketplace based on their profile details as well as marketplace metadata. This is the principle behind the ML-based marketplace \textit{curation} at Hired.

Candidate curation is a differentiator for Hired and leads to a better overall experience for recruiters (more likely to see relevant quality candidates with less noise than traditional inbound hiring platforms) as well as for candidates (higher chance of receiving interest from recruiters). It also entails developing and maintaining robust ML models capable of providing decisions in real-time. A highly demanded feature is the ability to provide actionable feedback to candidates with how to modify their profiles in order to improve their classifier scores and hence, their chance of being approved on the marketplace. 

Actionable feedback can be attained by generating \textit{counterfactuals}, which describe alternative scenarios. For instance, a profile with modifications to a candidate's salary or skills can be viewed as a "counterfactual" to the original profile. In this case, the actionable feedback would amount to the differences or modifications contained in the counterfactual compared with the original profile. In addition, counterfactuals are also useful for improving model interpretability, for instance, by highlighting potential biases and irregular predictions of a model (e.g., does the classifier's score change if a candidate's gender or race are modified?).

To ensure that the feedback is actionable, the counterfactual needs to be \textit{realistic} (e.g., no negative years of experience), require reasonable and few changes (i.e. \textit{sparse}), make sure it helps to \textit{achieve the desired outcome} (i.e., improve the classifier score), and should be able to be generated within \textit{real-time latency} constraints. Several counterfactual generation approaches have been proposed \cite{Wachter2017-jr, Van_Looveren2019-hr} but are limited in terms of realism and computational latency. A newly proposed CounteRGAN method \cite{Nemirovsky2020-te} was shown to be able to overcome these limitations by utilizing Generative Adversarial Networks \cite{Goodfellow2014-wf}.

In this work, we demonstrate how we are able to provide real-time actionable feedback to candidates by leveraging the CounteRGAN on Hired's dataset and production ML classifier. Our experiments show how this method is able to overcome the realism and latency limitations of previous counterfactual generation techniques and be practically feasible for production environments such as hiring marketplaces. For clarity, we also provide two real profile examples which highlight the potential of this method. To the best of our knowledge, we are the first to apply GAN-based counterfactual generation methods to a hiring marketplace dataset.

\section{Related Work\label{sec:related_work}}
\noindent
In contrast to counterfactuals produced using adversarial perturbation techniques \cite{Goodfellow2014-yo, Su2017-pk}, that aim solely at confusing a target classifier regardless of realism, counterfactuals as explanations for ML predictors were introduced by \cite{Wachter2017-jr}. In this work, counterfactual search was presented as a minimization problem approached utilizing gradient descent with regularization to account for sparsity. Though the resulting counterfactuals are generally of the desired target class, they are often unrealistic and very slow to compute. Several approaches have since been proposed to improve counterfactual realism. These include using a graph-based density approach \cite{Poyiadzi2019-qi}, autoencoders \cite{Dhurandhar2018-hk}, and class prototypes \cite{Van_Looveren2019-hr} to help influence the counterfactual search towards particular regions of the feature space. These approaches, however, are mainly aimed at differentiable models and are still extremely limited in terms of computational latency.

In order to provide real-time actionable feedback to candidates on the Hired marketplace, it is imperative that the counterfactuals be realistic, actionable, achieve the desired outcome, and computable within real-time latency constraints. Recently, a novel counterfactual generation method, termed CounteRGAN, was introduced in \cite{Nemirovsky2020-te}. It builds upon Generative Adversarial Networks \cite{Goodfellow2014-wf} to produce meaningful counterfactuals in real-time. This method, described in greater detail below, is able to overcome the limitations of previously mentioned approaches and allows for the capability of providing actionable feedback to be viable in many domains and industries.

\section{Approach using CounteRGAN \label{sec:approach}}
\noindent 
\begin{figure}[t]
\centering
\includegraphics[width=0.7\columnwidth]{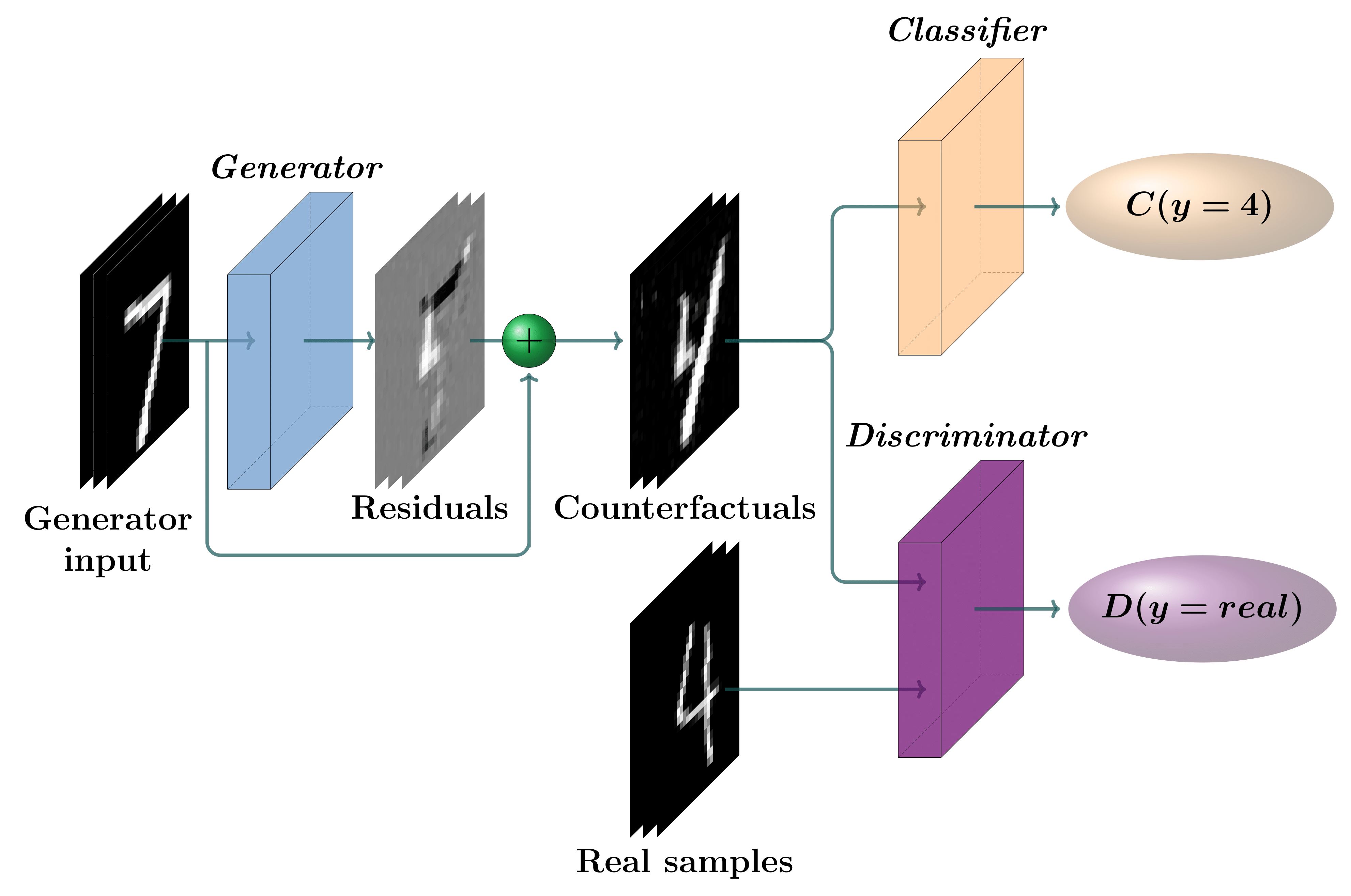}
\caption{CounteRGAN method on an example from MNIST \cite{Nemirovsky2020-te}. Three neural networks are used, a generator trained to output residuals, a discriminator trained to distinguish realistic data, and a target classifier. The example shows a generator outputting residuals that, when added with the input, produces realistic counterfactual images of a "4".}
\label{fig:countergan_arch}
\end{figure}

\noindent This section describes how we apply a GAN, in particular the CounteRGAN method, to generate counterfactuals capable of providing actionable feedback. GANs generally pit two artificial neural network models, termed a generator and discriminator, in an adversarial minimax game. During training the generator is trained to produce synthetic data that confuse the discriminator while the discriminator is trained to distinguish real data from synthetic data. The CounteRGAN method, shown in Figure \ref{fig:countergan_arch}, is a specialized GAN which relies on a residual generator $G$, a discriminator $D$, and a fixed classifier $C_t$ for a target class $t$. The formulation of the value function is given in Equation \ref{eq_countergan} below.

\begin{equation}
\label{eq_countergan}
 \mathcal{V}_\text{CounterRGAN}(G, D, C, t)=\mathbb{E}_{x \sim p_{\mathrm {data}}}\big[\log D(x)
    +  \log \left(1-D(x+G(x))\right)
  +\log \left(1-C_t(x+G(x))\right)\big] 
\end{equation}

where both the generator and discriminator use inputs samples from the same probability distribution $p_{\mathrm {data}}$ and we add a regularization term $ \lambda\mathbb E_{x \sim p_{\mathrm {data}}} \left\Vert G(x)\right\Vert_2^2 $ to control proximity.

For our purposes, we use Hired's curation ML classifier as the target classifier $C$ and candidate profiles from historical marketplace data as samples of $p_{\mathrm{data}}$. During training, we provide real candidate profiles to the generator as input and train it iteratively via gradient descent to minimize the value function (\ref{eq_countergan}) and produce counterfactuals that achieve high classifier scores while also appearing realistic to the discriminator. The discriminator is trained to maximize the value function and hence distinguish real candidate profiles from those synthesized by the generator. 

It is important to note that not all of the fields of a candidate's profile are mutable nor are their values all continuous. To account for immutable features such as a candidate's years of experience or education degrees, we force all the generated counterfactuals to have zero value changes to these features. To handle features that take on discrete values (e.g., "has a Ph.D. degree?", "headline word count"), we found that simply rounding the values of these features contained within the counterfactual produced favorable results.
\section{Experiments \label{sec:experiments}}
\noindent 

\renewcommand{\arraystretch}{1.35}
\begin{table*}[!ht]
\small
\centering
\begin{tabular}{l|c||c|c|c}
{} Metric &      Formula &               RGD &                    CSGP &              CounteRGAN \\
\hline
$\downarrow$ Realism &   $\left\Vert \mathrm{AE}(x_{\mathrm{cf}})- x_{\mathrm{cf}}\right\Vert_2^2$ &        1.08 $\pm$ 0.02 &  \textbf{0.69 $\pm$ 0.02} &  \textbf{0.68 $\pm$ 0.02} \\
$\uparrow$ Prediction gain     &  $C(x_{\mathrm{cf}}) - C(x)$ & \textbf{0.42 $\pm$ 0.02} &           0.07 $\pm$ 0.01 &           0.08 $\pm$ 0.01 \\
$\downarrow$ Actionability     &     $\left\Vert x_{\mathrm{cf}} - x\right\Vert_1$   &           1.31 $\pm$ 0.03 &  \textbf{0.16 $\pm$ 0.02} &           0.45 $\pm$ 0.02 \\
$\downarrow$ Latency (ms)        &  - &      1,696.65 $\pm$ 2.54 &        7,713.98 $\pm$ 9.82 &  \textbf{1.57 $\pm$ 0.02} \\
$\downarrow$ Batch latency (s)   &  - &               1,028.17 &                 4,674.67 &           \textbf{0.03} \\
\end{tabular}
\caption{Test data results (mean and 95 \% confidence interval). The arrows indicate whether larger $\uparrow$ or lower $\downarrow$ values are better, and the best results are in bold. The second column describes the formulas for the corresponding metrics, with $x_{\mathrm{cf}}$ being the suggested counterfactual and $x$ the input data sample for which a counterfactual is sought. The CounteRGAN method achieves impressive results and vastly outperforms other methods in terms of latency needed for real-time production systems.}
\label{table:metrics}
\end{table*}

\noindent We conduct our experiments using historical data sampled from candidates in the San Francisco market who registered on Hired from October 2019 to May 2020, consisting of 3,029 candidate profile samples, 43 \% of which have positive labels. We also use an 80-20 random split for the training and test data accordingly (results shown are based on performance on the test set).

The dataset contains 33 features that describe the candidate's profile as well as marketplace metadata at the time of submission. Of those, 27 are considered immutable (e.g., marketplace metadata, education, years of experience) and 6 are mutable (shown in the first column of Table \ref{table:examples}). The mutable features such as \textit{expected salary} and \textit{candidate headline word count} are rounded to the nearest integer or the nearest multiple of 5,000 in the case of salary. The \textit{experience relevance score}, \textit{verified years of experience}, and \textit{skills popularity score} are continuous-valued and are based on custom functions, the definitions of which fall outside the scope of this work.

We compare the CounteRGAN approach \cite{Nemirovsky2020-te} with the Regularized Gradient Descent (RGD) \cite{Wachter2017-jr} and Counterfactual Search Guided by Prototypes (CSGD) \cite{Van_Looveren2019-hr} methods described earlier in Section \ref{sec:related_work}. We optimize the corresponding parameters based on typical values used in the corresponding papers, slightly optimized through manual exploration. The methods are implemented using TensorFlow, and are run on a Tesla T4 GPU provided by Google Colab\footnote{https://colab.research.google.com/}. The ML classifier is based on a neural network production model used at Hired. The accuracy of this classifier on the test set is 76.4 \%.

To evaluate the relative performance of the different methods, we follow the four desirable properties of counterfactual generation defined in \cite{Nemirovsky2020-te}. \textit{Prediction gain}, measures the the difference between the classifier's prediction on the counterfactual and the input data sample. \textit{Realism} is based on the reconstruction error of generated counterfactuals using a denoising auto-encoder. \textit{Actionability} uses the $L1$ distance between the generated counterfactual and the input data sample. \textit{Latency} is the total execution time needed to compute one counterfactual and \textit{batch latency} is based on computing counterfactuals across the whole test dataset.

\begin{table*}[hbt!]
\scriptsize
\centering
\begin{tabular}{l||c|ccc||c|ccc}
 &  \multicolumn{3}{c}{$\qquad\qquad\qquad$\textbf{First example}} &  &  \multicolumn{4}{c}{\textbf{Second example}} \\
{} Feature &  Init. values &        RGD &     CSGP &  CounteRGAN & Init. values &        RGD &     CSGP &  CounteRGAN\\
\hline
Expected salary                     &       165,000 &  + 185,000 &  + 5,000 &    + 15,000 &  180,000 &  +5,000 &  0 &    +60,000  \\
Candidate headline word count             &            4 &       +7 &     0 &        + 5  & 4 &     +1 &  0 &        +8\\
Experience relevance score       &            0.72 &       +0.22 &     +0.06 &       + 0.46 &    0.53 &     +2.47 &  +0.06 &        +0.22  \\
Work experience avg word count     &           77.00 &     -18.42 &     0.00 &       +19.13 &      31.00 &    -5.81 &  0.00 &        +5.64  \\
Verified years of\ experience             &           10.83 &      -0.26 &     0.00 &       -4.71  &    18.83 &    -0.01 &  0.00 &        +1.09  \\
Skills popularity score        &            0.63 &       +2.67 &     +0.39 &        +0.76   &     0.94 &     +2.25 &  +0.07 &       -0.11 \\
\hline
\textit{Classifier prediction score}                &            \textit{0.35} &       \textit{0.97} &     \textit{0.50} &        \textit{0.72}     &            \textit{0.45} &      \textit{0.97} &  \textit{0.51} &        \textit{0.62}  \\
\end{tabular}
\caption{Examples of actionable feedback provided by the different methods for two real candidate profiles. The last row shows the classifier score for the initial candidate profile, and for the same profile after applying the suggested feedback. The CounteRGAN method is able to achieve realistic and actionable feedback capable of improving the classifier score over the 0.5 approval decision threshold.}
\label{table:examples}
\end{table*}

\subsection{Results}
\noindent
We present the results of our counterfactual generation experiments in table \ref{table:metrics}. RGD produces counterfactuals with the largest prediction gains, but are not realistic nor actionable. The CSGP approach produces more realistic and actionable counterfactuals with more modest prediction gains and with higher latency costs. Finally, the CounteRGAN method is able to produce realistic and actionable counterfactuals over a 1000x-30,000x faster than the other methods, capable of feasibly providing real-time actionable feedback in production settings.

Table \ref{table:examples} illustrates the type of actionable feedback and improvements in classifier prediction on two examples of real candidate profiles. In both cases, all methods were able to suggest feedback that actually results in the classifier approving the candidates (assuming a decision threshold of 0.5). In the first example, the RGD method suggests more than doubling the expected salary, which is not a realistic change. The CSGP method suggests minimal changes that increase the classifier score by 0.15, and the CounteRGAN suggests slightly larger updates to the input features but achieves a larger prediction gain (0.47). A similar pattern is observed for the second candidate. The feedback provided using the CounteRGAN method is not only actionable and of significant utility to the candidate, but is also able to be provided in real-time once the candidate submits their profile for admission to the marketplace.

\section{Conclusion \label{sec:conclusion}}
\noindent 
In this work, we have explored how hiring marketplaces such as Hired rely on ML classifiers for enhancing the marketplace quality and experience of candidates and recruiters via curation. This leads to a demand for providing actionable feedback to candidates to improve their profiles and chances of being approved onto the marketplace. We introduced the application of GANs for producing actionable feedback on hiring marketplaces such as Hired and illustrated its potential on production data, including two detailed examples using real candidate profiles. Our experiments demonstrate how a GAN-based approach is able to significantly outperform two other state-of-the-art methods (including over 1000x latency gains) to produce real-time actionable feedback within the constraints needed for production systems in industries such as hiring marketplaces.

\bibliography{industry_paper_refs.bib}

\bibliographystyle{unsrt} 
\end{document}